# From Planes to Corners: Multi-Purpose Primitive Detection in Unorganized 3D Point Clouds

Christiane Sommer[1]    Yumin Sun[1]    Leonidas Guibas[2]    Daniel Cremers[1]    Tolga Birdal[2]

*Abstract*—We propose a new method for segmentation-free joint estimation of orthogonal planes, their intersection lines, relationship graph and corners lying at the intersection of three orthogonal planes. Such unified scene exploration under orthogonality allows for multitudes of applications such as semantic plane detection or local and global scan alignment, which in turn can aid robot localization or grasping tasks. Our two-stage pipeline involves a rough yet joint estimation of orthogonal planes followed by a subsequent joint refinement of plane parameters respecting their orthogonality relations. We form a graph of these primitives, paving the way to the extraction of further reliable features: lines and corners. Our experiments demonstrate the validity of our approach in numerous scenarios from wall detection to 6D tracking, both on synthetic and real data.

## I. INTRODUCTION

**O**UR everyday environments are composed of a large number of man-made structures, that are constructed after careful computer aided design (CAD). As a result, they involve a large portion of simple geometric primitive forms.

Many of those primitives are planar, being either parallel or orthogonal to each other [1]. This renders the issue of discovering perpendicularity relationships in 3D a vital task for low level vision algorithms. In this paper we first propose a geometric voting-driven method to jointly detect pairs of orthogonal planes in oriented 3D point clouds, without explicitly resorting to segmentation or plane-grouping. We introduce a new local parameterization for orthogonal plane pairs. This allows us to cast votes in only a 2D local accumulator space, making our algorithm more efficient than hypothesis validation, as used in RANSAC-based approaches. Our approach is more reliable in detecting orthogonality than the standard "detect-then-build-graph" approach, since orthogonality in our case is directly deduced from data rather than from intermediate results (such as plane parameters).

We only cast one vote per point pair, which is significantly less computation than one inlier check on the whole point

Manuscript received: September 10, 2019; Revised December 06, 2019; Accepted January 05, 2020.

This paper was recommended for publication by Editor Cesar Cadena upon evaluation of the Associate Editor and Reviewers' comments. This work was partially supported by the ERC Consolidator Grant "3D Reloaded" and by Siemens AG.

[1]Christiane Sommer, Yumin Sun and Daniel Cremers are with the Computer Vision Group, TU Munich, Germany {sommerc, suny, cremers}@in.tum.de

[2]Leonidas Guibas and Tolga Birdal are with the Geometric Computing Group, Stanford University, CA, USA guibas@cs.stanford.edu, tbirdal@stanford.edu

Digital Object Identifier (DOI): see top of this page.

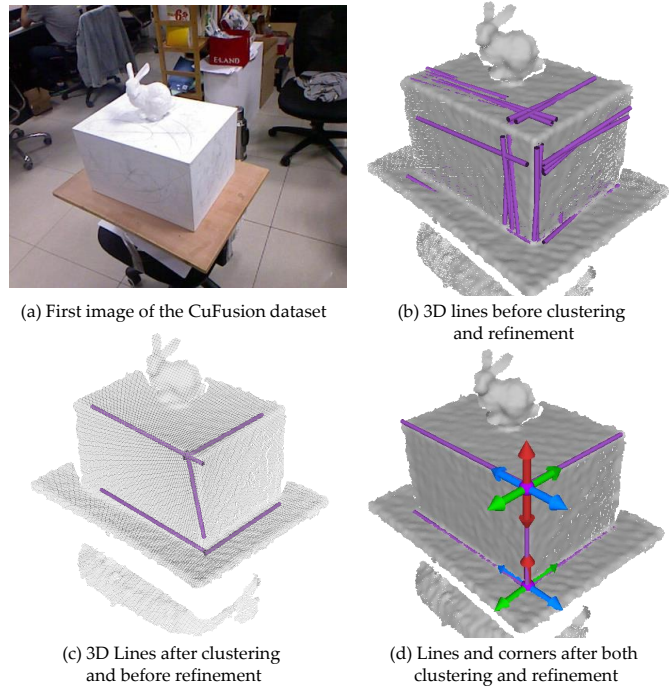

(a) First image of the CuFusion dataset

(b) 3D lines before clustering and refinement

(c) 3D Lines after clustering and before refinement

(d) Lines and corners after both clustering and refinement

Fig. 1: Steps of our algorithm on CuFusion dataset [2]. We can simultaneously detect orthogonal planes and their intersection lines (b,c), compute the orthogonal relation graph and use it to extract reliable corners with 6D local reference frames (d). Step (c) is intentionally stippled as the unoptimized lines fall behind the surface.

cloud per pair, as done in RANSAC [3]. Unlike region-growing [4], our algorithm can detect orthogonal pairs under occlusion, where planes can disconnect. The voting is remarkably similar to Hough transform of lines [5], [6] and extraction of intersection lines is achieved at no additional cost. A designated clustering follows the voting stage and subsequently, we build a relation graph out of the detected planes, where an edge depicts orthogonality between two planes (nodes). Thanks to this graph structure, we can significantly reduce the dimensionality of the joint parameter estimation problem. We then propose a novel, softly-constrained orthogonal refinement loss using this compact re-parameterization, to optimize for the alignment of planes to their support. Finally, we add the next layer of abstraction, in which by detecting triangles in the plane graph, we can arrive at *virtual* corners decorated with local reference frames (LRFs) directly computed from robustly fitted planes. The virtual points found on the intersection of three-planes are also highly accurate and can thus be used for tracking and ICP registration [7] as we will show. Our method is very efficient and can handle large datasets.

Overall, our algorithm that jointly extracts primitives at different abstraction layers (from planes to corners) contributes in the following (see Fig. 2 for an overview):

1) A novel scheme to vote for orthogonal plane pairs (and thus, their lines of intersection) without segmentation;
2) An efficient minimization scheme for constrained refinement over the reduced graph;
3) A 6D corner extractor where a corner is composed of a 3D location and an LRF.

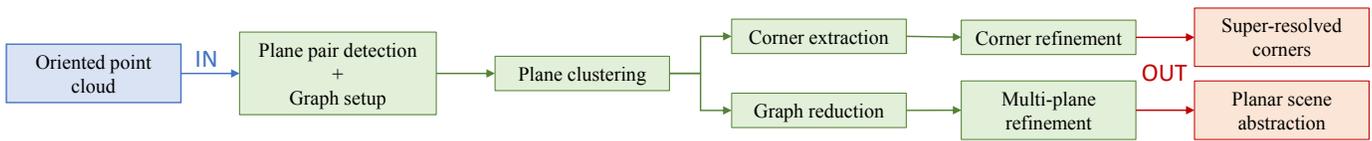

Fig. 2: Overview of our method: the detected plane pairs can serve for both corner extraction and planar scene abstraction.

## II. Related Work

We briefly review the literature concerned with extraction of 3D planar structures. For further details and references, we point the reader to a recent, extensive review on the general primitive detection from 3D data [8].

*a) Plane Detection:* While detecting 3D planar structures is harder and rather ill posed in 2D domain [9], [10], the problem is well studied in the 3D domain. Borrmann *et al.* use an enhanced Hough accumulator to vote for 3D planes [11]. Yang and Förstner [12] proposed a RANSAC based scheme, which is widely applied in different computer vision tasks. They use minimum description length (MDL) to deal with several competing hypotheses. Schnabel *et al.* [3] generalized the RANSAC based approach to detect different primitives in point clouds, including planes. Deschaud *et al.* [4] as well as Feng *et al.* [13] used filtered normals and voxel growing, a 3D analogous of region-growing, to devise a fast and accurate split-and-merge scheme. Further studies incorporated post-clustering and outlier elimination steps to robustify the pipelines [14], [15], [16], [17]. Drost and Ilic [18] introduced a Local Hough Voting scheme to retrieve multiple planes without segmentation, thanks to the use of point pair features. Similar to [19], [20], their algorithm addresses the discovery of other primitives such as spheres and cylinders but not pairs of orthogonal planes. It is also worth mentioning that using planes along with other primitives in a joint manner to approximate the objects has been tackled by many [18], [21], [22], [23].

*b) Orthogonality in Action:* Besides the Manhattan World reconstruction, a direct and widely accepted application of orthogonality to 3D data is SLAM (simultaneous localization and mapping). Many studies used the orthogonal planes as constraints in SLAM [24], [25], [26], [27], [28], [29], [30], or to aid robotic navigation [31], [32]. The common approach is to formulate SLAM to account for the orthogonality and directly use it in the pipeline. These works do not explicitly address the detection of the orthogonalities though.

The methods which consider problems similar to ours are [33], [34], [18]. In [33], Garcia *et al.* develop a box recognition algorithm. Many works in this family use triplets of points to define a plane. For the case of oriented point sets, this is an over-parameterization. Drost and Ilic [18] used point pair features for primitive detection. While their method is similar to ours, they only detect single primitive instances, without relations inbetween them. Jiang and Xiao [34] detect cuboids in images, but they do not on unstructured 3D data, as we do in this paper. Analogous to us, GlobFit [21] and Oesau *et al.* [35] use geometric regularization terms and relation graphs to position their primitives, but in contrast to us, these methods split the detection and relation graph building stage. Furthermore, they are designed for clean settings.

## III. Detection of Orthogonal Plane Pairs

Our method is separated into two stages: *detection* and *refinement*. The former extracts simplified *point pair features* [36], [37], [38] from the data, shows how to define orthogonality and devises a novel voting/clustering scheme for discovering orthogonal plane hypotheses. The latter simultaneously optimizes for all the parameters in the plane graph. Finally, the intersection points of the triangles in this plane graph lead to accurate corners.

*a) Orthogonal Point Pair Features:* The input to our method is a point set $\{\mathbf{x}_i\} \subset \mathbb{R}^3$ together with normals $\{\mathbf{n}_i\} \subset \mathbb{S}^2$. Note that the normals do not need to be consistently oriented, so if no normals are given, we can easily compute them by fitting planes to local neighborhoods. We parameterize a 3D plane $\mathbf{P}$ by a point and its normal. To characterize the orthogonal planes, we will speak of a pair of points which constitute the minimal set for defining an orthogonal plane pair (OPP). Imagine a pair of points $\mathbf{x}_1, \mathbf{x}_2$ with normals $\mathbf{n}_1, \mathbf{n}_2$, as shown in Fig. 3. Let $\mathbf{d} \in \mathbb{R}^3$ be the vector joining two points, i.e. $\mathbf{d} = \mathbf{x}_1 - \mathbf{x}_2$. If each of the two points lies on a plane, the condition for the two planes to yield an orthogonal configuration is

$$\angle(\mathbf{n}_1, \mathbf{n}_2) = \pi \,/\, 2 \,. \tag{1}$$

This can easily be re-written in terms of the scalar product $\mathbf{n}_1 \cdot \mathbf{n}_2$, allowing for an efficient computation. Yet, the data used in real life, e.g. from RGB-D sensors, never obey strict equality constraints. Hence, we introduce a noise threshold, maintaining certain tolerance: $|\mathbf{n}_1 \cdot \mathbf{n}_2| < \sin \delta_n$, where $\delta_n$ trades off noise tolerance vs accuracy. To make sure the two planes intersect at a meaningful point, one can further introduce a distance constraint: $\|\mathbf{d}\| < \tau_d$, where $\tau_d \in \mathbb{R}$ is a threshold.

Having all pair relations, we could now define the used point pair features (PPF) similar to [36], [18]. Since we do not need the actual angles, but can define the two constraints in terms of scalar products, we introduce simplified "features", that do not need trigonometry operations:

$$F(\mathbf{x}_1, \mathbf{x}_2) = \left(\mathbf{n}_1 \cdot \mathbf{n}_2, \mathbf{n}_1 \cdot \mathbf{d}, \mathbf{n}_2 \cdot \mathbf{d}, \|\mathbf{d}\|\right) \tag{2}$$

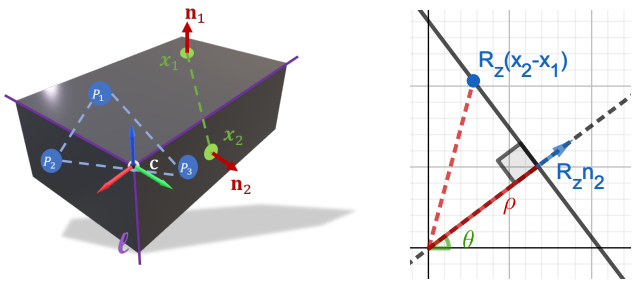

Fig. 3: **(left)** The concave geometric configuration that we are interested in. We jointly detect orthogonal planes and 3D lines, extract their relationship graph and obtain the corners. **(right)** The voting parameters $(\theta, \rho)$ shown in the 2D Cartesian system. $R_z$ is chosen such that $R_z \mathbf{n}_1 = \mathbf{e}_z$.

It will become clear in the voting section why we also keep the second and third components of $F$.

*b) Local Hough Voting:* Given a point pair, the definition of orthogonal planes is immediate – the normals of two points uniquely define the orthogonal planes. However, finding the best candidate requires care if the scene is cluttered and occluded. The trivial option is to perform RANSAC [39], where random point pairs are tested for the satisfaction of the orthogonal plane constraints $|\mathbf{n}_i \cdot \mathbf{n}_j| < \sin \delta_n$. While this is straightforward, it needs an inlier search at each step, making the whole procedure time consuming for large point sets. Local Hough voting, as in [18], circumvents this problem.

Thus, we evaluate the aforementioned constraints and create a voting table, similar to [18]. We sample the scene for $N$ reference points $\{\mathbf{x}_i\}$, each of which is paired with a maximum of $K$ other points in a $\tau_d$-neighborhood of $\mathbf{x}_i$ to compute the PPF. Each pair which satisfies the constraint casts a vote for the parameterization of the orthogonal planes to a local accumulator initiated per each reference point $\mathbf{x}_i$. While [18] uses such local voting for the detection of individual geometric primitives (planes, spheres, cylinders), we propose to port this idea to the detection of geometric relations inbetween planar primitives.

Once an oriented pair of points is found to be an OPP, the reference point $\{\mathbf{x}_1, \mathbf{n}_1\}$ defines the first plane. The orthogonal counterpart $\{\mathbf{x}_2, \mathbf{n}_2\}$, can freely rotate around the normal of the reference and is also free to slide orthogonally on this infinite reference plane. All such transformations result in the same PPF. Thus, we must resolve two degrees of freedom. We represent the second plane in 2D polar space, with respect to the reference: $(\theta, \rho)$. $\theta$ denotes the normal direction (which, being parallel to the reference plane, only has one degree of freedom), and $\rho$ is the orthogonal distance from the intersection line to the point of reference $\mathbf{x}_1$. The vote $(\theta, \rho)$ can be cast in 2D space, by transforming the point pair to the origin, and aligning $\mathbf{n}_1$ with the $z$-axis using a matrix $R_z$ as in Fig. 3. In analogy to Hough Transform of lines [5], [6], the variables of the local voting space read:

$$\theta = \arctan 2 \left( (R_z \mathbf{n}_2)_y, (R_z \mathbf{n}_2)_x \right) \tag{3}$$

$$\rho = \mathbf{n}_2 \cdot (\mathbf{x}_1 - \mathbf{x}_2) = \mathbf{n}_2 \cdot \mathbf{d} = F_3 \tag{4}$$

Voting is performed locally for each reference point, resulting in $\{\theta_1, ..., \theta_N\}$ and $\{\rho_1, ..., \rho_N\}$. For each reference point, two pairs describing the same set of orthogonal planes vote for the same $\theta$ and $\rho$. The voting also requires quantization of this local reference frame, which can be chosen reasonably depending on the problem size. The parameters $(\theta, \rho)$ with the maximal vote are taken to represent the most likely OPP and stored for each reference point if the vote count exceeds a count threshold $c_{max}$. This is important to make sure that very noisy reference points will not get accepted, and thus have an implicit noise handling. Note that this approach is semi-global and can recover the parameters even under severe occlusion [36].

In order to ensure that the reference point actually lies on a plane, we additionally track the number of paired points which are co-planar with $\mathbf{x}_1$, i.e. $|F_1| > \cos \delta_n$ and $|F_2|, |F_3| < F_4 \sin \delta_n$ and only insert a plane pair into the list of candidates if this number exceeds $c_{max}$.

*c) Clustering and Graphical Representation:* Rough detection results in an OPP hypothesis per selected reference point, giving rise to a pool of solutions which are to be clustered and merged. To this end, we use a disjoint forest clustering scheme [40] backed by a union-find structure. Planes are compared by computing the distance of each plane's reference point to the other plane.

In order to store all of the retrieved planes and their orthogonality relations, we choose a graph data structure: each plane is a vertex in the graph $\mathcal{G}$, and two planes $\mathbf{P}_i$ and $\mathbf{P}_j$ are connected by an edge $(i,j) \in \mathcal{E}$ if they intersect and are orthogonal. Special structures in the graph translate to special plane configurations, e.g. a triangle represents a corner in the point cloud and is endowed with the LRF composed of the normals of the triplet of orthogonal planes surrounding it. This can for instance be used as a 6D feature for tracking or scan registration. Note that our graph structure is similar to the one proposed in GlobFit [21], but with the difference that our graph is built during detection, whereas GlobFit separates the detection, relation extraction and graph building stages.

## IV. REFINEMENT OF ORTHOGONALITY PRIMITIVES

Due to sampling of the scene, quantization of the voting space, noise and artifacts, the orthogonal fitting obtained up to this point is only a rough estimate of the real pose of planes. Even though for certain applications this might well be sufficient, a refinement is still crucial for applications demanding accuracy. For that purpose, the most straightforward approach is a modified ICP-like non-linear optimization procedure, in which the distance from the points to the orthogonal planes are jointly minimized. While this has been done before in a very simple, unconstrained setting [7], we show how to use such modified ICP for joint plane refinement that respects the inter-plane geometric constraints – *first* for corners, which we parameterize efficiently in $\mathbb{R}^3 \times SO(3)$, and *second* for a multiplane setting, where we show how graph reduction can strictly enforce parallelity. We take advantage of the closed form expression of point-to-plane distances in order to avoid the costly nearest neighbor search. This way, we achieve a highly efficient method.

*a) Corner Refinement:* As mentioned earlier, we can retrieve corners in the given point cloud by finding triangles

in the plane graph $\mathcal{G}$. A corner found on the intersection of three orthogonal planes has six degrees of freedom, which can be used for tracking and scan registration. We formulate the objective function for corner refinement as:

$$E(\mathcal{X}, \{\mathbf{P}_1, \mathbf{P}_2, \mathbf{P}_3\}) = \sum_i \min_{k=1,2,3} r(\mathbf{x}_i, \mathbf{P}_k)^2 , \quad (5)$$

where $\mathbf{P}_1$, $\mathbf{P}_2$ and $\mathbf{P}_3$ denote the mutually orthogonal planes, $\mathcal{X} = \{\mathbf{x}_i\}$ is the point cloud and $r(\mathbf{x}, \mathbf{P}) = \mathbf{n} \cdot \mathbf{x} + d$ is the point-to-plane distance. Without further constraints, this energy has no orthogonality-preserving nature. In order to model this constraint, while still efficiently parameterizing the energy in Eq. (5), we rewrite our triplet as a tuple of three orthogonal normals $R := (\mathbf{n}_1 \quad \mathbf{n}_2 \quad \mathbf{n}_3)^\top$. The remaining three parameters to fully characterize the corner are the distances $d_1$, $d_2$, $d_3$ of the planes from the origin. Thus the corner refinement energy in Eq. (5) becomes

$$E(\mathcal{X}, \mathbf{d}, R) = \sum_i \min_{k=1,2,3} (R\mathbf{x}_i + \mathbf{d})_k^2 \quad (6)$$

with $\mathbf{d} = (d_1, d_2, d_3)^\top \in \mathbb{R}^3$ and $R \in SO(3)$. This parameterization also endows the corner with an LRF that is unique up to sign flips. Note that the initialization of Sec. III does automatically ensure the orthogonality of plane pairs. Yet, when it comes to the mutually orthogonal triplets this is no longer the case. To ensure, we make use of the fact that the frame composed of the triplet normals has a diffeomorphic mapping to $SO(3)$ and project $R$ onto $SO(3)$. It is important to make sure that $R$ is not a reflection but a rotation: we switch the order of $\mathbf{n}_2$ and $\mathbf{n}_3$ if $det(R) < 0$. $R$ is 3D and can be re-parameterized using twist-coordinates [41] for efficient optimization on the $SO(3)$ manifold without resorting to costly constrained optimization.

In a real-world setting, if we want to use corners for tracking or alignment, we need to only use data points $\mathbf{x}$ which are close to the 3D corner $\mathbf{c}$, to avoid outliers. $\mathbf{c}$ is given by $-(d_1\mathbf{n}_1 + d_2\mathbf{n}_2 + d_3\mathbf{n}_3)$. Thus, we define a subset

$$\mathcal{X}_\mathbf{c} := \{\mathbf{x} \in \mathcal{X} : \|\mathbf{x} - \mathbf{c}\| < \epsilon\} \subset \mathcal{X} \quad (7)$$

on which we perform the optimization, i.e. we instead minimize $E(\mathcal{X}_\mathbf{c}, \{\mathbf{P}_1, \mathbf{P}_2, \mathbf{P}_3\})$. Strictly speaking, $\mathcal{X}_\mathbf{c}$ implicitly depends on the plane parameters. However, in practice, we use the initial estimate $\mathbf{c}$ to select the point set $\mathcal{X}_\mathbf{c}$, and then keep $\mathcal{X}_\mathbf{c}$ fixed.

*b) Multi-plane Refinement and Parameter Reduction:*
For geometry refinement, it is important that we can refine all planes in a scene jointly. The unconstrained energy for this scenario is

$$E(\mathcal{X}, \{\mathbf{P}_k\}) = \sum_i \min_k r(\mathbf{x}_i, \mathbf{P}_k)^2 \quad (8)$$

Again, this energy totally lacks the notion of orthogonality between planes. Furthermore, in the case that the graph $\mathcal{G}$ contains vertex groups of a specific structure, we can in addition to orthogonality deduce which planes are parallel [21], which is also not being taking into account in Eq. (8). We address the two constraint types (orthogonal and parallel) differently:

first, we re-structure our graph $\mathcal{G}$ by combining parallel planes into one node, where each node is endowed with a list of distances $\{d_{kl}\}$. Then, we write the energy as

$$E'(\mathcal{X}, \{(\mathbf{n}_k, \{d_{kl}\})\}) = \sum_i \min_{k,l} (\mathbf{n}_k \cdot \mathbf{x}_i + d_{kl})^2 . \quad (9)$$

This way, the normal vector for each set of parallel planes needs to be optimized only once, significantly reducing the number of unknowns. Thus, the parallelity constraint is enforced by re-parameterization. Second, we add an orthogonality regularizer $W$ to $E'$:

$$W(\mathcal{G}) = \sum_{(k,k') \in \mathcal{E}} (\mathbf{n}_k \cdot \mathbf{n}_{k'})^2 \quad (10)$$

with the edge set $\mathcal{E}$, resulting in the regularized energy:

$$E_W(\mathcal{X}, \mathcal{G}) = E'(\mathcal{X}, \mathcal{G}) + \lambda W(\mathcal{G}) . \quad (11)$$

Since reprojection to the feasible set of parameters is not straightforward, $\lambda$ needs to be large enough to implicitly enforce the orthogonality between planes. In order to avoid further constraints on unit length of $\mathbf{n}_i$, we use the on-manifold optimization following the $\mathbb{S}^2$-parameterization given in [23] to achieve $\|\mathbf{n}_k\| = 1$ for all $k$. Note that our refinement is more principled than the iterative approximation of the constraint satisfaction in GlobFit [21] and relies less on heuristics: we avoid inputting a fixed points-to-planes assignment to the refinement. Rather, the cost function in Eq. (8) by construction re-assigns points to their closest plane in each iteration of a minimization scheme. This procedure is more tolerant to wrong assignment in the detection phase and thus more accurate by construction. This also explains why GlobFit expects good initialization and clean input or else quickly gets stuck in a local minimum.

For robustness, we add an M-estimator $\Phi$ to the data term $E'$. In terms of computational complexity, we keep costs for the point-to-plane assignment low in two ways: (1) we only compute the point-to-plane distance if the angle between plane normal and plane normal is below a certain threshold $\epsilon_n$, and (2) for each $l$, we sort the according $d_{kl}$ such that the time to find the $\arg\min_k$ is halved.

*c) Application: Corner-assisted ICP Registration:* The 3D position $\mathbf{c}$ of a corner can be anywhere in 3D space – in particular, it does not need to exactly co-incide with a data point. Thus, the accuracy of $\mathbf{c}$ depends only very weakly on the sampling density of the given point set, and much more on its noise level. This way, we can see the corners as *super-resolved* key points in our point cloud. We use this fact in order to improve ICP-based registration of two point clouds: on the one hand, we can use the corners for coarse alignment of two scans. On the other, the refinement, which is typically done via ICP, can also take advantage of the corners: For a set of corners $\{\mathbf{c}_k\}$ in the target point cloud and the corresponding set $\{\mathbf{c}'_k\}$ in the coarsely aligned source point cloud, we know that the $SE(3)$-transform bringing the two into correspondence has to satisfy:

$$(R, \mathbf{t}) \in \arg\min_{(R,\mathbf{t})} \sum_k \|R\mathbf{c}'_k + \mathbf{t} - \mathbf{c}_k\|^2 =: A_c \subset SE(3) ,$$

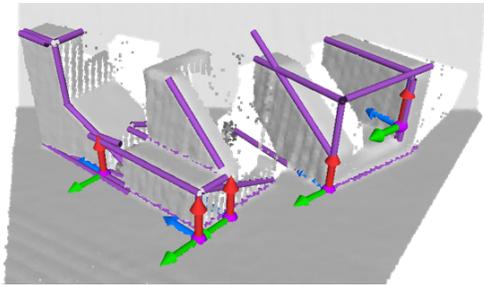

| Modality | Algorithm | Pr. | Rec. | #Cor. | Noise | Miss |
|---|---|---|---|---|---|---|
| O-Planes | AHC [13] + R.G. | **0.97** | 0.81 | **9.77** | **0.33** | 2.23 |
| | Schnabel [3] + R.G. | 0.85 | 0.59 | 7.13 | 1.33 | 4.87 |
| | Ours | 0.88 | **0.87** | 9.76 | 1.38 | **1.83** |
| Lines | AHC [13] + R.G. | 0.74 | 0.65 | 8.53 | 3.63 | 4.67 |
| | Schnabel [3] + R.G. | 0.57 | 0.34 | 4.43 | 5.20 | 8.77 |
| | Ours | **0.77** | **0.73** | **8.86** | **2.90** | **3.97** |

Fig. 4: Orthogonal plane detection on the Orthogonal SegComp/ABW [42] dataset. We visualize the extracted line and corner primitives on the left. The table on the right reports the precision (Pr), recall (Rec), number of correct detections (#Cor) as well as false positives (Noise) and false negatives (Miss). *O-Planes* refers to the results of detecting planes that have orthogonal pairs. This corresponds to the vertices in our relation graph (R.G.). *Lines* refers to evaluating the edges of the graph, corresponding to the orthogonal planes. This one evaluates the performance of 3D line extraction.

i.e. the transformation needs to align corresponding corners. We explicitly choose to only align the 3D positions of corners, since the rotation precision we obtain in an $\epsilon$-neighborhood (which we use to find the corner position) easily becomes too low if scans are to be aligned globally. Note that the set $A_c$ can have more than only one element. $\dim A_c$ depends on the number of corner correspondences:

- If there are at least three corners that do not all lie on one line, $\dim A_c = 0$ and the unique minimizer $(R, \mathbf{t})$ is given by the Kabsch algorithm [43].
- If two corners are present, or more corners that lie on one single line, $\dim A_c = 1$ and the elements in $A_c$ differ by rotation with an angle $\alpha$ about that line.
- If there is only one corner, $\dim A_c = 3$ with

$$A_c = \{(R, \mathbf{t}) : \mathbf{t} = \mathbf{c}_1 - R\mathbf{c}_1'\} \cong SO(3) . \quad (12)$$

In order to align the two point sets $\{\mathbf{x}_i\}$ and $\{\mathbf{x}_i'\}$, the ICP algorithm minimizes the cost

$$E_{\text{ICP}}(R, \mathbf{t}) = \sum_i \left( (R\mathbf{x}_i' + \mathbf{t} - \mathbf{x}_{j(i)}) \cdot \mathbf{n}_{j(i)} \right)^2 \quad (13)$$

with $\mathbf{x}_{j(i)}$ being the nearest neighbor of $R\mathbf{x}_i' + \mathbf{t}$. Having super-resolved corner positions, instead of minimizing $E_{\text{ICP}}$ over all $SE(3)$, we constrain the problem to $(R, \mathbf{t}) \in A_c$. This gives rise to new energies $E_{\text{ICP}}^{1c}(R)$ and $E_{\text{ICP}}^{2c}(\alpha)$, with lower-dimensional domains. The advantages of this dimensionality reduction are two-fold: not only does lower-dimensional optimization converge faster, but also, less data points are needed for the optimizer to converge to a minimum at all. Briefly stated, we can use the high accuracy of corner positions to either completely omit ICP ($\geq 3$ corners), or to constrain the ICP problem such that a minimum can be found with less computation. We will also demonstrate this in the experiments section.

## V. EXPERIMENTAL EVALUATION

*a) Datasets:* In order to demonstrate the broad applicability of our proposed method, we evaluate our multi-purpose primitives on different datasets including SceneNN [44], ICL-NUIM [45], Cu3D [2] and Redwood [46]. It is noteworthy that for the task of primitive detection and discovery there are not many designated datasets. Due to the availability of ground truth (GT) segmentation and comparison metrics, we choose to augment the seminal SegComp dataset [42] with orthogonal planes, resulting in the augmented *Orthogonal SegComp* (O-SegComp). SegComp is a database of 30 scenes taken with a laser scanner. To create O-SegComp, we first fit planes robustly to the GT segmentation to extract GT plane parameters. We then build the relation graph and keep only those planes that have orthogonal counterparts. Ground truth data for O-SegComp thus consists of a subset of the SegComp planes, together with orthogonality information.

*b) Implementation Details:* We use the Ceres solver (ceres-solver.org) for energy minimization in all experiments. In particular, we locally parameterize $R \in SO(3)$ via Sophus (strasdat.github.io/Sophus) for the corner refinement in Eq. (6). For the multi-plane graph refinement in Eq. (11), we use a local parameterization of $\mathbb{S}^2$ [23] to represent the unit length normals. Prior to operation, we downsample large point sets to ensure spatial uniformity [47]. In particular, we sample the points that are at least $d_{\min}$ apart and average the samples reducing the noise, whenever present. To preserve the efficacy, we apply a coarse-to-fine refinement, where the optimizer uses a hierarchy of samplings gradually increasing the resolution and hence enhancing and accelerating convergence. We compute the surface normals, which don't need to be consistently oriented, by local plane fits. Our code, together with some pseudocode for easier understanding can be found here:

https://github.com/c-sommer/orthogonal-planes

*c) Choice of Parameters:* Starting from the parameters given in [36], [18], we experimented with different settings to find the optimal trade-off between speed and accuracy, arriving at the following: We are using a set of 500–2000 reference points (low if speed is critical, high for higher accuracy) and pair them with about 250 points in a $\tau_d$-neighborhood, where $\tau_d = 1$m. The normal threshold $\delta_n$ is set to $20°$ in all experiments, and the voting bin sizes for $\theta$ and $\rho$ are $10°$ and 8cm, respectively. We accept the bin with the highest vote $c_{\max}$ as plane pair candidate if $c_{\max} > 4$. The parameter $d_{\min}$ for downsampling the point cloud is chosen adaptively, based on size and shape of the point set.

### A. Quantitative Results

*a) Detection of Planes and Intersection Lines:* We begin by evaluating the ability of our algorithm in extracting planes that belong to an orthogonal pair. We use O-SegComp for this and we report precision, recall, number of correct detections (true positives), noise (false positives) and misses (false

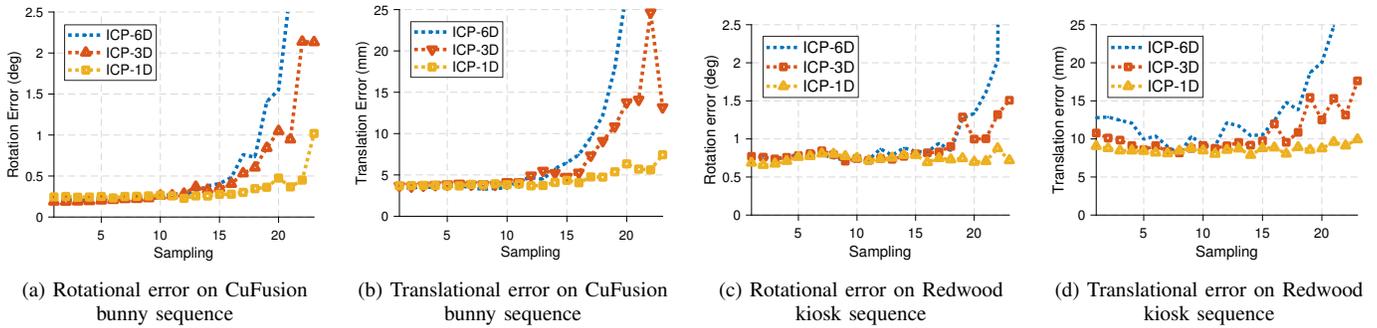

(a) Rotational error on CuFusion bunny sequence   (b) Translational error on CuFusion bunny sequence   (c) Rotational error on Redwood kiosk sequence   (d) Translational error on Redwood kiosk sequence negatives). Depicted as *O-Planes* in Fig. 4, our results are comparable to those of the state of the art.

Fig. 5: Rotational and translational errors in corner assisted ICP on the CuFusion [2] and Redwood-Kiosk [46] datasets.

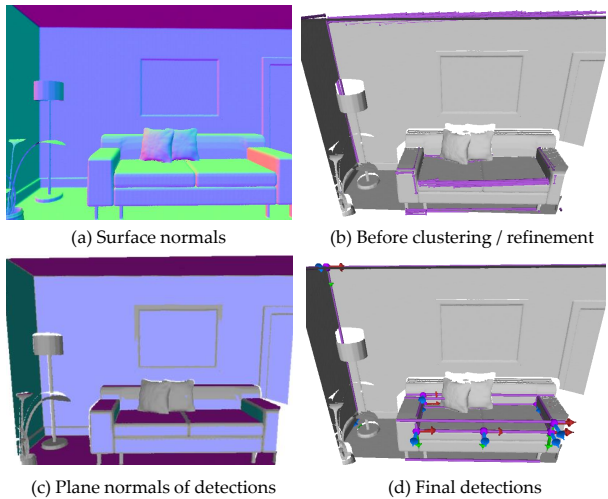

(a) Surface normals   (b) Before clustering / refinement

(c) Plane normals of detections   (d) Final detections

Fig. 6: Detection of orthogonality primitives on an ICL-NUIM scene.

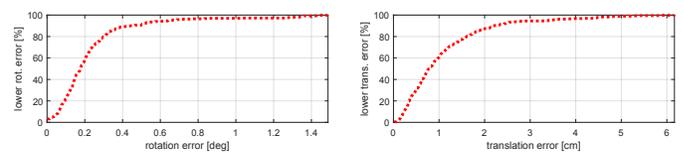

Fig. 7: Cumulative plots of rotational and translational RPE on the ICL-NUIM livingroom sequence after single-corner alignment. For an error $x$, we plot the percentage of successful alignments with an error below $x$. We find that more than 90% of matches have a rotation error below 0.5 degrees, and close to 90% of matches have less than 2cm translation error.

To evaluate our second layer primitives, we intersect the planes of the Orthogonal SegComp, yielding the ground truth intersection lines. We then use an analogous evaluation metric to the plane case. We consider a line to be correctly detected if the angle it makes with the ground truth match is less than $10°$. We report the result in Fig. 4. Note that there is a drop in the performance as opposed to the plane case. This is because missing a single plane yields an entire set of missing lines. Nevertheless, our approach still maintains a recall of 73% with a precision of 77%, better than using plane detectors of AHC [13] and Schnabel [3], together with subsequent relation graph building. We would like to emphasize that these results suggest that joint detection of plane parameters and orthogonality is a promising research direction, as it leads to better detection of orthogonality compared to the standard "detect-then-build-graph" approach. In the same figure, we also show the quality of our detection.

*b) Corner-Assisted ICP Registration:* As described earlier, sets of corresponding corners can be used to constrain the domain of the registration energy. To this end, we augment a standard implementation of the ICP algorithm by taking into account the corners we detect in a pair of scans. This experiment is to be understood as a proof of principle: we show that reliably extracted corners can improve a standard tracking/registration algorithm by comparing the baseline (no corners) to different corner-assisted modalities. In order to demonstrate the effect of super-resolved corners on ICP registration, we sample random scenes from the CuFusion bunny sequence [2] as well as Redwood Kiosk sequence [46] and run a pairwise registration. We plot the median of the relative pose error (RPE) against the downsampling factor for scans that are temporally about 0.33 and 0.1 seconds apart for CuFusion and Redwood sequences respectively. Fig. 5 shows the errors attained at the rotational and translational components. We consider the relative poses obtained from the DVO RGB-D odometry [48] as ground truth. This algorithm also uses the RGB information, so it is consistently more accurate, which justifies our GT choice. The figure shows that, while being comparable to the RPE of full 6DoF-ICP for dense sampling, the RPE of corner-assisted low-DoF tracking increases much slower and thus ICP-1D shows more stability even for high downsampling factors. Constraining the possible solution set of ICP to a lower-dimensional set thus in particular pays off if only sparse data is available or if the sampling density for any reason must be held low. In Fig. 9, we additionally show that both the time and the number of iterations until convergence go down for lower DoF ICP, which is expected, since the space of parameters that we optimize over is smaller.

*c) Corner Alignment on ICL-NUIM:* To asses the quality of 6D alignment of scans by matching detected corners, we do pairwise alignment of a subset of all frames by sampling each 15th frame of the sequence. Matching is considered successful if the overlap of the two scans after alignment is sufficiently large. For all pairs of successfully matched scans, we compute the relative pose error. Fig. 6 (left) shows the detected primitives whereas Fig. 7 reports the cumulative plots for the rotational and translational RPE components. Like the corner-assisted ICP, this experiment serves as a proof of concept. We are not aware of any other global registration algorithm that registers scans only based on one single 6D

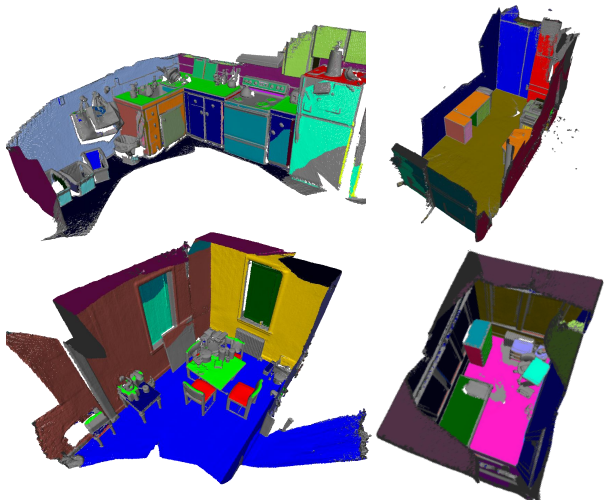

Fig. 8: 3D segmentation on SceneNN [44] by labeling the planes that are found to have at least one orthogonal "partner". Note, the chairs in the lower-left are assigned a common plane, typically hard to achieve by region-growing [13].

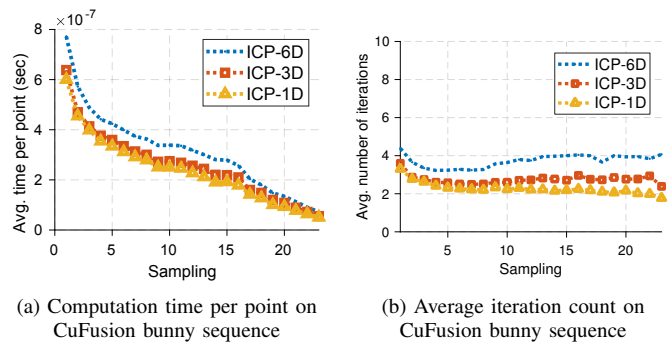

(a) Computation time per point on CuFusion bunny sequence

(b) Average iteration count on CuFusion bunny sequence

Fig. 9: Statistics in corner assisted ICP registration on the CuFusion [2] bunny sequence.

feature.

*d) Runtimes:* On a desktop PC with an Intel Xeon CPU @3.5 GHz a single-threaded implementation needs about 10ms for voting and candidate extraction, and between 80-300ms for graph refinement, depending on the scene complexity. Thus, we are orders of magnitudes faster than GlobFit, which can (even without re-assigning points to planes during iterations) take 3-5min for a point cloud of comparable size. Other methods [3], [13] do not refine the inter-plane configurations, which renders a fair time comparison hard. The efficient RANSAC [3] implementation in CGAL (cgal.org) needs about 70ms for extracting planes on the full cloud, and roughly 7ms for 2000 points. After plane extraction, a subsequent refinement is necessary to obtain accurate planes. AHC [13] can extract planes at more than 35Hz from depth data at VGA resolution, but it needs structured point clouds, whereas we can work with any type of point set. Further, AHC does not set planes in context to one another, which is the main point of our proposed method.

### B. Qualitative Evaluations

*a) Orthogonal Plane Segmentation in Real Data:* In Fig. 8 we show the success of the detected orthogonal plane pairs in semantically summarizing scenes where perpendicular plane configurations are dominant. This is typically the case for our man-made indoor environments. Thus, we choose the SceneNN [44] dataset and run our detector. As we do not explicitly result in a segmentation map, but directly compute plane parameters, we assign points to planes by considering a closeness-threshold and a normal coherence.

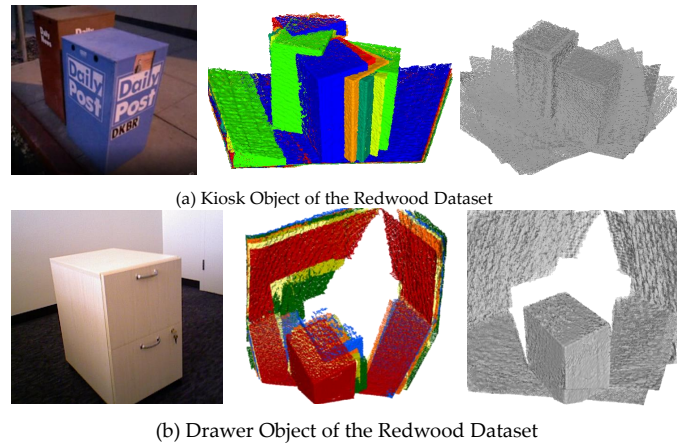

(a) Kiosk Object of the Redwood Dataset

(b) Drawer Object of the Redwood Dataset

Fig. 10: Our accurate corners are used to align multiple scans. We qualitatively evaluate this 6D registration on the Redwood dataset [46]. For each subfigure, we show from left to right: (1) The RGB image (only used for visualization), (2) input scans overlayed on top of each other in the local coordinate frame of each camera, (3) scans shown after alignment to the frame of the first camera.

*b) 3D Reconstruction via Corner Alignment:* Our final experiment involves reconstruction via detection, where a collection of unordered scans are processed to estimate pairwise transformations. Typically, the desired relative poses are found by some form of descriptive matching be it global or local. In 3D, most descriptors suffer from the ambiguities in the LRF. At this point our 6D corners can be helpful providing principled and reliable means of registration. We illustrate this in Fig. 10, where a scan of Kiosk and a scan of Drawer from the Redwood dataset [46] are brought into a consistent global alignment by registering the LRF of the found corners. Just like for the pairwise corner alignment, we consider the matching successful if the overlap after transformation is sufficiently large.

## VI. CONCLUSION

We design a joint detection-refinement pipeline for orthogonal planes and higher-level primitives, such as lines and corners of intersection, on sparse or dense point sets. This is the first work incorporating semi-global PPFs into a local voting framework for this purpose. Our novel 2D local parametrization is sufficient to establish the full (5D) pose of orthogonal plane configurations. The method alleviates discretization artifacts from at least three of the parameters, while maintaining speed and accuracy. We can detect multiple orthogonal plane pairs and cluster them to describe the 3D geometry of the environment. Thanks to the optimization step, all the approximate orthogonal configurations detected in 3D point clouds can be refined up to machine precision and sensor noise yielding a very precise fit. In the future, we will extend our framework to even higher-level primitives like boxes etc. and use our orthogonal planes in SLAM.